\documentclass[10pt,twocolumn,letterpaper]{article}

\usepackage{iccv}
\usepackage{times}
\usepackage{epsfig}
\usepackage{graphicx}
\usepackage{amsmath}
\usepackage{amssymb}
\usepackage{multirow}
\usepackage{xcolor}
\usepackage{color, colortbl}
\definecolor{mygray}{gray}{.85}

\usepackage[breaklinks=true,bookmarks=false]{hyperref}

\iccvfinalcopy 

\def\iccvPaperID{****} 

\begin{document}

\title{TrajectoryFormer: 3D Object Tracking Transformer with Predictive \\Trajectory Hypotheses}
\author{Xuesong Chen\textsuperscript{1} \quad
Shaoshuai Shi\textsuperscript{2 }\thanks{Corresponding authors} \quad
Chao Zhang\textsuperscript{3}  \quad
Benjin Zhu\textsuperscript{1}  \quad
Qiang Wang\textsuperscript{3}  \\
Ka Chun Cheung\textsuperscript{4}  \quad
Simon See\textsuperscript{4} \quad
Hongsheng Li\textsuperscript{1, 5, 6 }\protect\footnotemark[1] \\
\protect \textsuperscript{1}MMLab, CUHK \quad
\protect \textsuperscript{2} Max Planck Institute for Informatics \quad 
\protect \textsuperscript{3} Samsung Telecommunication Research \quad \\
\protect \textsuperscript{4} NVIDIA AI Technology Center \quad 
\protect \textsuperscript{5} CPII \quad 
\protect \textsuperscript{6} Shanghai AI Laboratory \quad \\
{\tt\small \{chenxuesong@link, hsli@ee\}.cuhk.edu.hk, shaoshuaics@gmail.com}
}
\maketitle

\begin{abstract}
3D multi-object tracking (MOT) is vital for many applications
including autonomous driving vehicles and service robots.
With the commonly used tracking-by-detection paradigm, 3D MOT has made important progress in recent years. However, these methods only use the detection boxes of the current frame to obtain trajectory-box association results, which makes it impossible for the tracker to recover objects missed by the detector. In this paper,  we present TrajectoryFormer, a novel point-cloud-based 3D MOT framework.  To recover the missed object by detector, we generates multiple trajectory hypotheses with hybrid candidate boxes, including  temporally predicted boxes and current-frame detection boxes, for trajectory-box association. The predicted boxes can propagate object's history trajectory information to the current frame and thus the network can tolerate short-term miss detection of the tracked objects.  We combine long-term object motion feature and short-term object appearance feature to create per-hypothesis feature embedding, which reduces the computational overhead for spatial-temporal encoding.
Additionally, we introduce a Global-Local Interaction Module to conduct information interaction among all hypotheses and models their spatial relations, leading to accurate estimation of hypotheses. Our TrajectoryFormer achieves state-of-the-art performance on the Waymo 3D MOT benchmarks. Code is available at \textcolor{red}{https://github.com/poodarchu/EFG}.
   
\end{abstract}

\section{Introduction}

3D multi-object tracking (MOT) is an essential and critical task in the fields of autonomous driving and robotics. It plays a vital role in enabling systems to accurately perceive their surrounding dynamic environment and make appropriate responses. 
Among the various sensors used in autonomous driving, LiDAR-based systems have emerged as a popular choice because they can capture accurate and detailed 3D information of the environment, enabling more precise object detection and tracking. Therefore, 3D MOT based on LiDAR point clouds shows great potential to improve the safety and efficiency of autonomous vehicles.

Tracking-by-detection is a popular paradigm that has demonstrated excellent performance on the 3D MOT task~\cite{yin2021center,simpletrack,immortaltracker,benbarka2021score,poschmann2020factor}. Previous methods, such as CenterPoint~\cite{yin2021center} and SimpleTrack~\cite{simpletrack}, rely on heuristic rules to associate objects across frames. These methods use manually designed affinity metrics such as distance, intersection over union (IoU), and GIoU to match a history trajectory with a current detection box based on their positional relationship. However, these heuristic rules are not robust as they cannot be trained and different categories may prefer different association metrics~\cite{simpletrack}. Moreover, these methods only consider pair-wise position relationships between boxes, without considering comprehensive global context information, which often results in low-quality trajectories. 

Other methods have attempted to enhance 3D MOT by modeling the spatial context among different boxes. PolarMOT~\cite{kim2022polarmot} adopts Graph Neural Network (GNN) to establish the spatial-temporal relationship between trajectories and different boxes, followed by edge classification to conduct association.  Similarly, InterTrack~\cite{willes2022intertrack} employs attention mechanisms to interact between trajectories and all boxes, generating the affinity matrix for association. These methods generally leverage global context information, resulting in improved tracking performance compared to heuristic methods.
However, they still only rely on detection boxes for associating with existing trajectories, which limits the recall rate when the detector misses objects.
Thus, incorporating additional box candidates for association has great potential to improve the recall and performance of 3D MOT.

To overcome the limitations of existing approaches, we present TrajectoryFormer, a point-cloud-based 3D MOT framework. Our framework generates multiple trajectory hypotheses with hybrid candidate boxes, enabling robust tracking of challenging objects. It employs a per-hypothesis feature encoding module and a cross-hypothesis feature interaction module to learn representative features for selecting the best hypotheses.
The per-hypothesis feature encoding module encodes both the appearance and motion information of each hypothesis, whereas the feature interaction module captures the contextual relationship among all hypotheses. 
By leveraging multiple hypotheses and contextual information, TrajectoryFormer can enhance tracking performance in challenging scenarios with limited overhead. 

Specifically, our framework first generates multiple trajectory hypotheses for each existing trajectory using two types of association candidate boxes:  temporally predicted boxes and current frame detection boxes. 
Unlike existing approaches that only consider detection boxes at the current frame, we design a small motion prediction network that generate predicted boxes for several future frames of each history trajectory.
This allows us to generate multiple trajectory hypotheses for an object by linking its history trajectory with both temporally predicted boxes (generated by its motion prediction at different past time steps) and current frame detection boxes (matched by nearest distance of box centers). 
Such a strategy enables the network to recover objects missed by the detector at the current moment and provides additional association options that can help correct trajectory errors caused by low-quality detection boxes.

After generating multiple trajectory hypotheses, TrajectoryFormer combines long-term object motion feature and short-term object appearance feature to create per-hypothesis feature embedding. 
More specifically, we adopt a PointNet-like~\cite{shi2022motion} neural network to encode the motion feature for each trajectory hypothesis via encoding its long-term sequence boxes, and a small transformer-based neural network on the cropped object points to encode its appearance feature.
Note that we only encode the object appearance feature based on short-term point clouds, since it not only requires very limited computational overhead but also avoids handling long-term object point variations.
The concatenation of two types of features that capture complementary information for each trajectory hypothesis creates the per-hypothesis feature embedding. This embedding enables the evaluation of each hypothesis quality and facilitates the modeling of relationships among multiple hypotheses.

To jointly consider the trajectory association across all objects, we introduce a global-local Interaction module that models spatial relations of all trajectory hypotheses.
It uses a transformer-based neural network to alternately conduct scene-level (\emph{e.g.}, all trajectory hypotheses within the scene) and ID-level (\emph{e.g.}, multiple trajectory hypotheses of each object) feature interactions on the hypotheses, leading to more accurate estimation of hypotheses. 
During inference, TrajectoryFormer selects the hypothesis with the highest confidence as the best association result for each object. The selected hypothesis is then refined using its extracted features to generate a more accurate trajectory.

In summary, our contributions are three-fold:
1) We propose TrajectoryFormer, a novel transformer-based 3D MOT tracking framework, which generates multiple trajectory hypotheses that incorporate both predicted and detected boxes to better track challenging objects.
2) To better encode each hypothesis, we incorporate both long-term trajectory motion features and short-term object appearance features. Additionally, the framework employs a global-local interaction module to model relationships among all hypotheses to adaptively determine the optimal trajectory-box association.
3) We demonstrate the effectiveness of our proposed approach through extensive experiments, and our framework achieves state-of-the-art results for 3D MOT on the challenging Waymo 3D tracking benchmark.

\begin{figure*}[t]
\centering
\includegraphics[width=0.99\textwidth ]{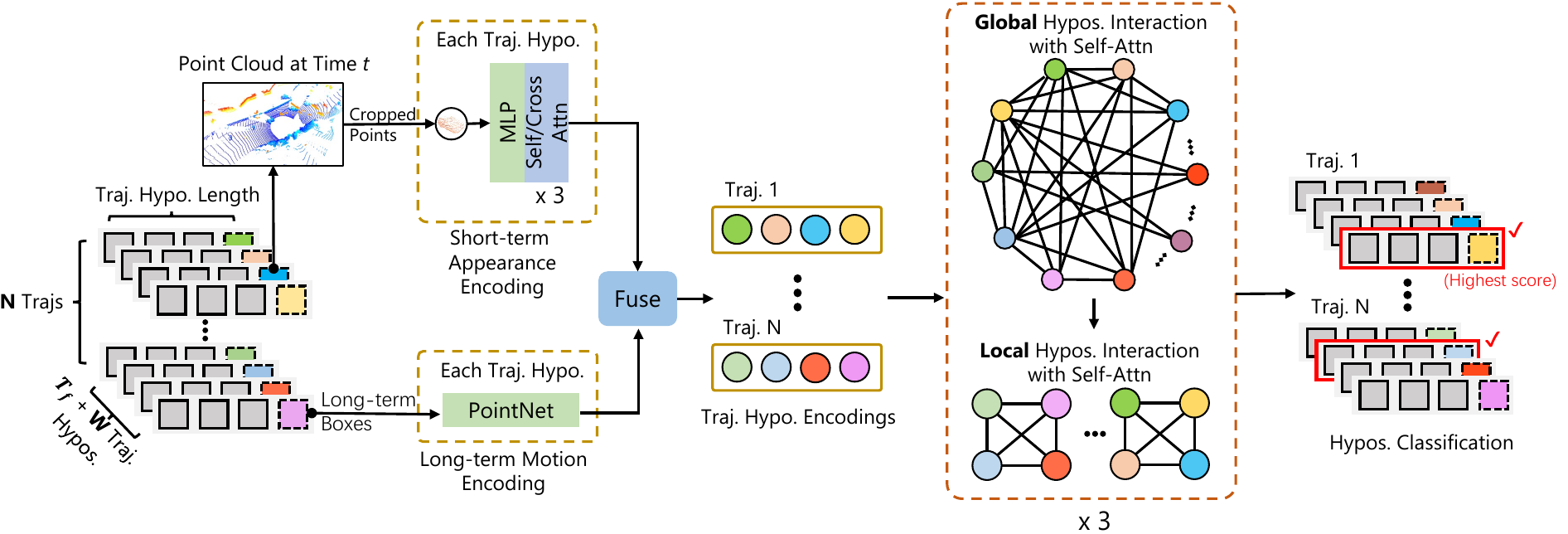}
\caption{The overall framework of the proposed TrajectoryFormer. Given $N$ history trajectories and the input point cloud, we first generate multiple trajectory hypotheses for each history trajectory by incorporating both $W$ detected boxes and $T_f$ temporally predicted boxes. Then a long-short hypothesis feature encoding module is used to encode the appearance and motion feature of each hypothesis. These hypothesis features are then further encoded via a global-local hypothesis interaction module to propagate information among these hypotheses. Finally, these features are utilized to predict the confidence of each hypothesis for selecting the best trajectory hypothesis.}
\vspace{-3mm}
\label{fig:pipeline}
\end{figure*}

\section{Related Work}

\subsection{3D Object Detection on Point Clouds}
Current methods of 3D detection on point cloud can be categorized into three groups: point-based, voxel-based, and point-voxel-based. The point-based methods~\cite{shi2019pointrcnn,qi2019deep,yang2019std} directly extract information from the original point clouds. These methods leverage operations like set abstraction~\cite{qi2017pointnet++} to capture spatial position features of the irregular 3D point clouds.
In contrast, voxel-based approaches convert the irregular points into regular 3D voxels. Therefore, voxel-based works~\cite{yan2018second,zhou2018voxelnet,yin2021center} can utilize 3D CNN to directly extract features of each voxel in 3D space.
Some high-efficiency methods~\cite{yang2018pixor,lang2019pointpillars,wang2020pillar,ge2020afdet} further reduce the height dimension of voxels, named pillar, and adopt a bird-eye view (BEV) representation to encode 3D features efficiently. 
Additionally, point-voxel-based methods aim to enhance the performance by leveraging the strengths of both point and voxel representations. 
By exploiting the two representations, some recent point-voxle-based works~\cite{shi2020pv_waymo,li2021lidar,shi2021pv} have achieved state-of-the-art detection results.

On the other hand,  some methods aim to exploit the benefit of multi-frame point cloud for better detection performance. Early methods employ a feature-based strategy to aggregate temporal features with 2D CNN~\cite{luo2018fast} or transformer-based architectures~\cite{9393615,9438625}. Recent works~\cite{hu2021afdetv2,sun2021rsn,yin2021center} have shown that a simple concatenation strategy of multi-frame points can significantly outperform the single-frame setting. 
Furthermore, MPPNet~\cite{chen2022mppnet} proposes to employ proxy point as a medium to handle information aggregation of long point clouds sequences.

\subsection{3D Multi-Object Tracking}

Benefiting from recent advancements in 3D object detection, the state-of-the-art 3D MOT algorithms have adopted the tracking-by-detection paradigm. A notable example is CenterPoint~\cite{yin2021center}, which proposes a simple approach that utilize objects' center distance as the association metric to link detection boxes across sequential frames. 
However, CenterPoint employ a constant velocity assumption to compensate for the motion displacement between different frames. This approach may exhibit less resilient to missing detections or curved motion trajectories.

Similar to the conception of optical flow~\cite{huang2022flowformer,shi2023videoflow,shi2023flowformer++}, several 3D MOT algorithms utilize Kalman Filters to estimate the location of tracked objects. AB3DMOT~\cite{abmot3d} serves as a foundational approach in this regard, where 3D Intersection-over-Union (IoU) is employed as the association metric for object tracking. 
Furthermore, Chiu \emph{et al.}~\cite{prob-mahabolis} propose an alternative approach by introducing the use of Mahalanobis distance as a replacement for 3D IoU to capture the uncertainty of the trajectories. Meanwhile, SimpleTrack~\cite{simpletrack} conducts an analysis of the different components of a tracking-by-detection pipeline and and provides suggestions for enhancing each component.
ImmortalTracker~\cite{immortaltracker} propose a simple tracking system that maintain tracklets for objects gone dark to solve the ID switch problem in 3D MOT.
SpOT~\cite{stearns2022spot} introduces a approach by developing the representation of tracked objects as sequences of time-stamped points and bounding boxes over a long temporal history. At each timestamp, SpOT improves the location estimates of tracked objects by utilizing encoded features from the maintained sequence of objects.

Some works exploit trajectory prediction to deal with occlusion problems in tracking or detection. 
FutureDet~\cite{peri2022forecasting} propose an end-to-end approach for detection
and motion forecasting based on LiDAR, which is capable of forecasting multiple-future trajectories via future detection. 
Quo-Vadis~\cite{dendorfer2022quo} utilizes trajectory prediction to solve long-term occlusions 
in single-camera tracking. Similarly, PF-Track~\cite{pang2023standing} maintains the object positions and enables re-association by integrating motion predictions to handle long-term occlusions multi-camera 3D MOT.

\section{TrajectoryFormer}

Existing state-of-the-art 3D MOT approaches~\cite{yin2021center,simpletrack,immortaltracker,stearns2022spot} generally adopt the tracking-by-detection paradigm, which utilizes the detected boxes at the current frame for trajectory-box association. 
Although these approaches have achieved excellent tracking performance, they may encounter difficulties when tracking challenging objects, such as occluded or distant objects, due to mis-detections or inaccurate localization caused by sparse object points. 
To address these limitations, we present an efficient framework, TrajectoryFormer, for 3D MOT in point cloud scenarios.  

Specifically, as shown in Fig.~\ref{fig:pipeline}, TrajectoryFormer generates a novel set of multiple trajectory hypotheses, which incorporate both current frame detection boxes and history-trajectory prediction boxes to better cover the potential moving patterns of tracked objects. 
In Sec.~\ref{sec:hypo_generation}, we first introduce the generation of multiple trajectory hypotheses. Next, in Sec.~\ref{sec:hypo_encoding}, we present the feature encoding of each trajectory hypothesis. In Sec.~\ref{sec:hypo_interaction}, we propose the global-local feature interaction module to propagate information among all the trajectory hypotheses and generate the final trajectories.
Finally, we introduce the losses of TrajectoryFormer in Sec.~\ref{sec:loss}. 

\subsection{Generation of Multiple Trajectory Hypotheses}\label{sec:hypo_generation}

Given $N$ existing history trajectories up to time $t-1$, $\{h_i^t\}_{i=1}^{N}$, 
state-of-the-art 3D MOT approaches~\cite{simpletrack,immortaltracker} generally associate each history trajectory with its nearest detection boxes at $t$ for extending the trajectories up to current time $t$.  However, this association strategy may fail in tracking some challenging objects if the detector misses the object at time $t$. 
To address this limitation, TrajectoryFormer is designed to generate \textit{multiple trajectory hypotheses} for each tracked object $h_i^t$ to better cover the potential moving patterns of each object. 
Unlike existing approaches that solely consider current frame detection boxes for trajectory-box association, 
each history trajectory $h_i^t$ is paired with hybrid candidate boxes at time $t$ to generate trajectory hypotheses, which include not only the current frame detection boxes but also the temporally predicted boxes based on the motion prediction of the history trajectory $h_i^t$.

\begin{figure}
    \centering
    \includegraphics[width=0.4\textwidth]{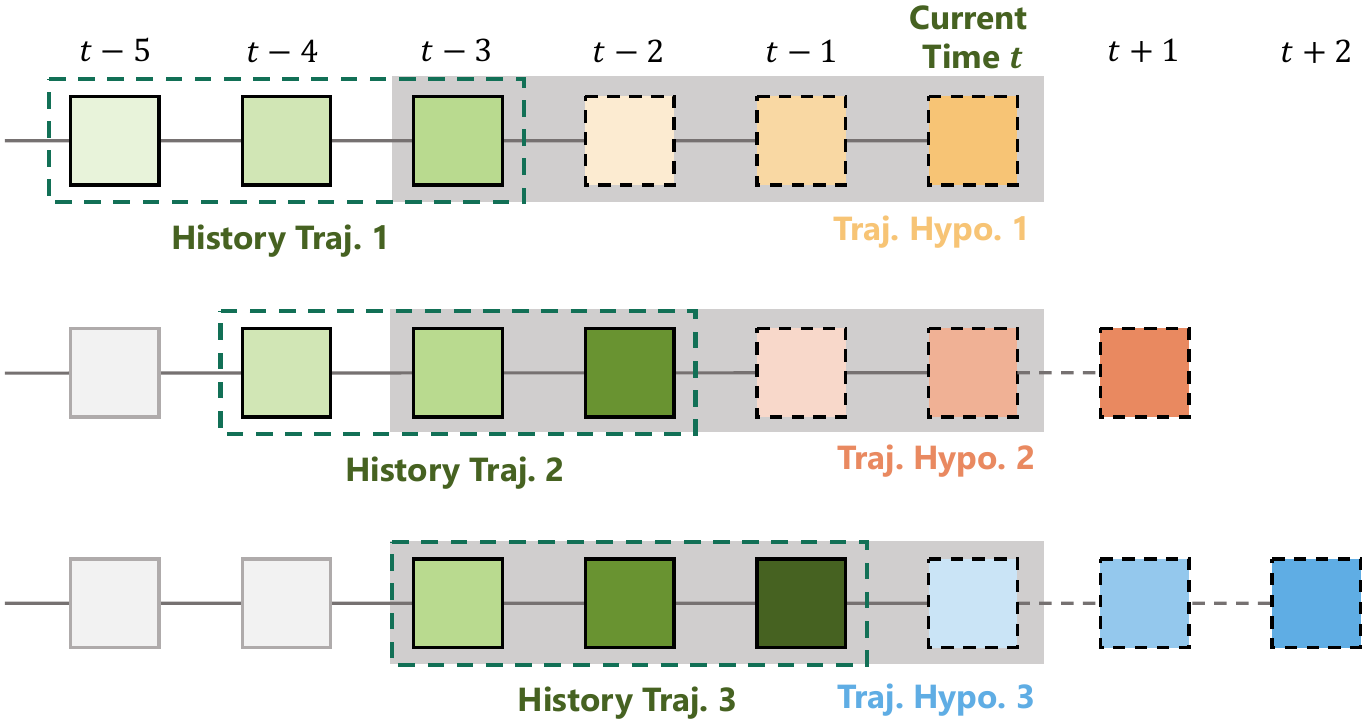}
    \caption{The illustration of the generation of multiple trajectory hypotheses at frame $t$ for a single history trajectory.}
    \label{fig:motion_prediction}
    \end{figure}

\noindent
\textbf{Motion Prediction of History Trajectories.}
To achieve this goal, we introduce a motion prediction network that encodes the historical trajectories of tracked objects to predict their future states. 
Specifically, we first reorganize the history trajectories $\{h_i^t\}_{i=1}^{N}$ as $H^t=\{\hat{h}_i^t \mid  \hat{h}_i^t\in \mathbb{R}^{T_h\times S}\}_{i=1}^{N} \in \mathbb{R}^{N\times T_h\times S}$. Note that $\hat{h}_i^t\in \mathbb{R}^{T_h\times S}$ is the cropped history trajectory of the $i$-th trajectory $h_i^t$ up to time $t-1$ with temporal length $T_h$, and $S$ denotes the number of state attributes at each frame, such as location, heading angle, velocity, and time encoding. We pad all-zero vectors to the history trajectories that are shorter than $T_h$. 

The motion features of each history trajectory are then encoded using a PointNet-like encoder as
\begin{align}\label{eq:point_encoding}
H_{g}^t &= \text{MaxPool}(\text{MLP}(H^t)),
\end{align}
where  $\text{MLP}(\cdot)$ is a multi-layer perception network transforming each $S$-dimensional history trajectory's state vector, followed by max-pooling over the temporal dimension to summarize all frames' features into $N$ history trajectory features $H_g^t \in \mathbb{R}^{N\times D}$. The trajectory features are then used as input to an MLP prediction head that predicts each object's future states as
\begin{align}\label{eq:point_encoding2}
H_p^t &= \text{MLP}(H_{g}^t),
\end{align}
where $H_p^t \in \mathbb{R}^{N\times T_f\times 3}$ is the set of predicted states at future $T_f$ frames for each history trajectory $h_i^t$ up to time $t-1$. $H_p^t$ can be reformulated as a set $H_p^t=\{p_i^t\}_{i=1}^{N}$, where $p_i^t\in \mathbb{R}^{T_f\times 3} $ indicates the predicted future states of the $i$-th trajectory starting at time $t$, and 3 represents the predicted 2D location and the heading angle at each time step.

\noindent
\textbf{Generation of Multiple Trajectory Hypotheses.}
With the predicted trajectory states from the past time steps, TrajectoryFormer generates multiple trajectory hypotheses for each tracked object by associating each history trajectory to each of its $T_f$ temporally predicted boxes and current-frame detected boxes. 

Specifically, as shown in Fig.~\ref{fig:motion_prediction}, for a given history trajectory $h_i$ up to time $t-1$, we collect its predicted states at time $t$ from the motion prediction results at the previous $T_f$ frames of the history trajectory, which can be represented as the set
\begin{align} 
\{p_i^{t-1}[1], \dots, p_i^{t-j}[j], \dots, p_i^{t-T_f}[T_f]\}\nonumber,
\end{align}
where $p_i^{t-j}[j]$ indicates the predicted state at current time $t$ by using the short clip of this history trajectory at time $t-j$. 
Note that we only predict the future position and heading angle of each trajectory and assume that the box dimension is unchanged for each history trajectory.
For the sake of simplicity, we consider $\{p_i^{t-j}[j]\}_{j=1}^{T_f}$ as the predicted boxes of the $i$-th history trajectory at current time $t$, which are utilized to associate with the $i$-th history trajectory to generate multiple trajectory hypotheses. We illustrate the generation of multiple trajectory hypotheses with temporally predicted boxes in Fig. \ref{fig:motion_prediction}, where the prediction length $T_f = 3$ and history length $T_h =3$.

In addition to the boxes temporally predicted from the past, each history trajectory $h_i^t$ is also associated with the $W$ detection boxes at the current frame, which are generated by a 3D detector and chosen as the nearest boxes to the trajectory.
We denote the associated detection boxes of the $i$-th history trajectory as $\{d_i^j\}_{j=1}^W$.

Given the generated predicted boxes and associated detection boxes, for each history trajectory $h_i^t$, we can obtain a set $\Omega$ consisting of $M=N\times(T_f+W)$ hypotheses as
\begin{align}\tiny 
    \Omega_p &= \{h_i^t\oplus p_i^{t-1}[1], \dots, h_i^t\oplus p_i^{t-T_f}[T_f]\}, \nonumber\\
    \Omega_d &= \{h_i^t\oplus d_i^1, \dots, h_i^t\oplus d_i^W \}, \\
    \Omega_{~} &= \Omega_p \cup \Omega_d, \nonumber
\end{align}
where $\oplus$ indicates linking the $i$-th history trajectory with a temporally predicted or detection box to generate a trajectory hypothesis.

These proposed multiple trajectory hypotheses strategy provides two key benefits. 
Firstly, it enables the recovery of objects that were not detected at time $t$ by propagating past times' temporal prediction boxes to time $t$, which creates trajectory hypotheses better tolerating short-term mis-detection of the tracked objects. 
Secondly, it provides more association candidates and can correct tracking errors in trajectories caused by low-quality detection boxes, since the 3D detector only uses limited temporal information in the point cloud sequence (\emph{e.g.}, 2-3 frames) and may produce low-quality detection boxes for challenging objects. 
In such cases, temporally predicted boxes might provide better trajectory hypotheses to improve the tracking quality.

\subsection{Long Short-Term Hypothesis Feature Encoding}\label{sec:hypo_encoding}

After obtaining multiple trajectory hypotheses, 
TrajectoryFormer adopts a long-short feature encoder to transform the trajectory hypotheses into the feature space, which involves encoding the long-term motion information and the short-term appearance of each trajectory hypothesis.

For long-term motion encoding, we employ a PointNet-like neural network that takes $M$ trajectory hypotheses' box sequence $\Omega_B \in \mathbb{R}^{M\times (T_h+1) \times 8}$ as input, where 8 means the number of boxes' properties (\emph{e.g.} 7-dim geometry and 1-dim time encoding), and outputs their motion features $E_m \in \mathbb{R}^{M \times D}$ as
\begin{align}
E_m = \text{MaxPool}(\text{MLP}(\Omega_B)).
\end{align}
The incorporation of such long-term motion information is crucial in differentiating hypotheses that exhibit similar appearance and location at the current time.

To reduce the computational cost and avoid handling long-term object point variations, we only encode the short-term appearance of each trajectory hypothesis. 
Specifically, we randomly sample $Y$ points by cropping the input short-term point cloud within the box at time $t$ of each trajectory hypothesis.  
We follow MPPNet~\cite{chen2022mppnet} to encode the box information to each cropped object point by computing the relative differences 
between each sampled point $p_i$ and 9 representative points  of the hypothesis box (8 corner and 1 center points). 
By appending an extra one-dimensional time offset embedding, the final point-wise appearance features of the $j$-th hypothesis of the $i$-th tracked object can be further encoded with an MLP network, which can be represented as $O_i^j\in \mathbb{R}^{Y\times D}$. 

Given these encoded point-wise features, we first utilize the self-attention mechanism to perform information interaction among all points and then adopt a cross-attention layer to obtain the aggregated embedding from $Y$ points as
\begin{align}
    \hat{O}_i^j &= \text{SelfAttn} ( Q(O_i^j), ~K(O_i^j), ~V(O_i^j) ) , \nonumber \\
    V_i^j &= \text{CrossAttn} ( Q(v), ~K(\hat{O}_i^j), ~V(\hat{O}_i^j) ) ,
\end{align}
where $i\in \{1, \cdots, N\}$ and $j\in \{1, \cdots, T_f+W\}$. $Q(\cdot), K(\cdot), V(\cdot)$ are linear projection layers to generate query, key, value features for the attention layers. $v \in \mathbb{R}^{1\times D}$ is zero-initialized learnable parameters to aggregate features from all the subsampled points of the $j$-th hypothesis of the $i$-th tracked object, which generates its final appearance feature as $V_i^j\in \mathbb{R}^{D}$. 

In practice, the self-attention and cross-attention operations are iteratively repeated for multiple rounds to update the query vector $v$ gradually. The final short-term appearance features of all $M$ hypotheses can be denoted as $E_{a} = \{ V_{i}^j\}_{i=1, j=1}^{N, T_f+W} \in \mathbb{R}^{M \times D}$.

Given the appearance embedding $E_a$ and motion embedding $E_m$, the long short-term embedding $E \in \mathbb{R}^{M \times D}$ of all $M$ hypotheses of trajectory $i$ is formed by concatenating the features with a one-hot class vector $C$ to distinguish their target category along the channel dimension as
\begin{align}
\label{eq:embed}
E = \text{MLP}(\text{Concat}(E_a, E_m, C)).
\end{align}

\subsection{Global-local Feature Interaction of Multiple Trajectory hypothesis}\label{sec:hypo_interaction}

The hypothesis features encode the appearance and historical motion information of each tracked object. However, it fails to consider the relationship between multiple trajectory hypotheses of the same tracked object and the interactions between all tracked objects in the same scene.

To properly model inter-hypothesis and inter-trajectory relations, we propose a Global-local Interaction Module to model the spatial relationships among all hypotheses of all tracked objects. 
Specifically, we  design a transformer with self-attention mechanism to propagate information between all trajectory hypotheses.
The interaction is performed alternatively between global and local contexts, as depicted in Fig.~\ref{fig:pipeline}. 
During global interaction, each hypothesis gathers information from all other hypotheses as
\begin{align}
\label{eq:global_attn}
 G = \text{SelfAttn} (Q(E), K(E), V(E))),
\end{align}
 which forms global-interacted embedding $G \in \mathbb{R}^{M\times D}$. 
On the other hand, local interaction emphasizes the interaction between different hypotheses of the same tracked object. Specifically, we use $G_{i}^{j}$ represents the globally-interacted $j$-th hypothesis embedding of the $i$-th tracked object, where $j=\{1, \dots, T_f+W\}$, $i=\{1, \dots, N\}$.  Therefore, the local interaction of $T_f+W$ hypotheses of the $i$-th tracked object can be expressed as
\begin{align}
\label{eq:local_attn}
 L_i^j = \text{SelfAttn} (Q(G^{j}_{i}), K(G^{j}_{i}), V(G^{j}_{i}))),
\end{align}
 which forms local-interacted embedding $L \in \mathbb{R}^{M\times D}$.

We alternately conduct the global and local interaction in the transformer for several times, which enables the representations of the hypotheses to incorporate both global and local contexts, allowing each hypothesis to gain a better understanding of the distribution of its neighboring objects. This module leads to improved association outcomes since the embedding of each hypothesis becomes more context-aware. 
After the interaction process, an MLP head is appended to generate a final probability score for each hypothesis' confidence evaluation, which is utilized for selecting the best trajectory hypothesis for of each tracked object. 

\subsection{Losses}\label{sec:loss}
The overall training loss contains two loss terms:
a confidence-score loss $\mathcal{L}_{\mathrm{conf}}$ and a bounding-box regression loss $\mathcal{L}_{\mathrm{reg}}$ as
\begin{equation}
\mathcal{L}= \mathcal{L}_{\mathrm{conf}} + \mathcal{L}_{\mathrm{reg}}.
\label{eq:loss}
\end{equation}
We adopt the binary cross entropy loss for $\mathcal{L}_{\mathrm{conf}}$, which is introduced to supervise the network to predict confidence-score of all trajectory hypotheses. 
For $\mathcal{L}_{\mathrm{reg}}$, we employ the same box regression loss in MPPNet~\cite{chen2022mppnet} to supervise box refinement, that is, to predict the residual of the position, shape and heading angle between hypothesis boxes and ground truth.
In addition, the simple motion prediction network is trained separately. We use the L1 loss to supervise the networks' prediction of future trajectory states, including center location and heading angle.

\section{Experiments}
In this section, we first outline our experimental setup, which includes the datasets, evaluation metrics, implementation details and life cycle management. Subsequently, we present comprehensive comparisons with state-of-the-art methods on the Waymo 3D MOT benchmarks. Finally, we provide a range of ablation studies and related analyses to investigate various design choices in our approach. 
\subsection{Dataset and Implementation Details}
\noindent
\textbf{Waymo Open Dataset.} The Waymo Open Dataset comprises 1150 sequences, with 798 training, 202 validation, and 150 testing sequences, and each of which contains 20 seconds of continuous driving data within the range of [-75m, 75m]. 
3D labels are provided for three classes, including vehicle, pedestrian and cyclist.

\noindent
\textbf{Nuscenes Dataset.} 
The nuScenes dataset is a large dataets that contains 1000 driving sequences and each sequence spans 20 seconds.
LiDAR data in nuScenes is provided at 20Hz but 3D labels are only given at 2Hz. We evaluate on the two most observed classes: car and pedestrian.

\noindent
\textbf{Evaluation Metrics.}
We adopt the official evaluation metrics as defined by the Waymo and nuScenes benchmarks for comparison.  
For Waymo, MOTA is employed as the primary evaluation metric, which involves three types of errors: false positives (FP), missing objects (Miss), and identity switches (IDS) at each timestamp. Furthermore, the evaluation performance is divided into two difficulty levels: LEVEL 1 and LEVEL 2. The former evaluates objects with more than five points, while the latter includes objects with at least one point. We use LEVEL 2 as the default performance setting. For nuScenes, we follow the official tracking protocol and use AMOTA as the main metric.

\begin{table*}[t]
\centering
\footnotesize
\begin{tabular}{l|cccccccccccc}
\hline
\multirow{2}{*}{Method}       & \multicolumn{4}{c}{Vehicle}                       & \multicolumn{4}{c}{Pedestrian}                    & \multicolumn{4}{c}{Cyclist}  \\
                              & \cellcolor{mygray}MOTA$\uparrow$ & FP\%$\downarrow$ & Miss\%$\downarrow$ & {IDS\%$\downarrow$} & \cellcolor{mygray}MOTA$\uparrow$ & FP\%$\downarrow$ & Miss\%$\downarrow$ & {IDS\%$\downarrow$} & \cellcolor{mygray}MOTA$\uparrow$ & FP\%$\downarrow$ & Miss\%$\downarrow$ & IDS\%$\downarrow$ \\ \hline
AB3DMOT~\cite{abmot3d}                       &\cellcolor{mygray}55.7 & -    & -      & {0.40}  & \cellcolor{mygray}52.2 & -    & -      & 2.74  & \cellcolor{mygray}-    & -    & -      & -     \\
CenterPoint~\cite{yin2021center}                   & \cellcolor{mygray}55.1 & 10.8 & 33.9   & {0.26}  & \cellcolor{mygray}54.9 & 10.0 & 34.0   & {1.13}  & \cellcolor{mygray}57.4 & 13.7 & 28.1   & 0.83  \\
SimpleTrack~\cite{simpletrack}                   & \cellcolor{mygray}56.1 & 10.4 & 33.4   & {0.08}  & \cellcolor{mygray}57.8 & 10.9 & 30.9   & {0.42}  & \cellcolor{mygray}56.9 & 11.6 & 30.9   & 0.56  \\
ImmotralTrack~\cite{immortaltracker}                 & \cellcolor{mygray}56.4 & 10.2 & 33.4   & {0.01}  & \cellcolor{mygray}58.2 & 11.3 & 30.5   & {0.26}  & \cellcolor{mygray}59.1 & 11.8 & 28.9   & 0.10  \\
SpOT~\cite{stearns2022spot}                          &\cellcolor{mygray} 55.7 & 11.0 & 33.2   & {0.18}  & \cellcolor{mygray}60.5 & 11.3 & 27.6   & {0.56}  &\cellcolor{mygray} -    & -    & -      & -     \\ \hline
Ours (CenterPoint) & \cellcolor{mygray}\textbf{59.7} & 11.7   & 28.4      & {0.19}     & \cellcolor{mygray}\textbf{61.0}    & 8.8    & 29.8  & {0.37}     & \cellcolor{mygray}\textbf{60.6}    & 13.0    & 25.6      & {0.70}      \\ 
Ours (MPPNet) & \cellcolor{mygray}\textbf{61.0} & 10.9   & 28.0      & {0.13}     & \cellcolor{mygray}\textbf{63.4}    & 11.6    & 24.6  & {0.40}     & \cellcolor{mygray}\textbf{63.5}    & 8.2    & 28.0      & {0.28}      \\ \hline
\end{tabular}
\caption{Tracking performance on the Waymo Open dataset \textit{validation} split. The employed detector of compared tracking methods, including SimpleTrack, ImmotralTrack and SpOT are all CenterPoint.}
\label{tab:waymo-val}
 \vspace{-3mm}
\end{table*}

\begin{table*}[!t]
\centering
\footnotesize
\begin{tabular}{l|cccccccccccc}
\hline
\multirow{2}{*}{Method}       & \multicolumn{4}{c}{Vehicle}                       & \multicolumn{4}{c}{Pedestrian}                     & \multicolumn{4}{c}{Cyclist}   \\
                              &\cellcolor{mygray} MOTA$\uparrow$ & FP\%$\downarrow$ & Miss\%$\downarrow$ & \multicolumn{1}{c}{IDS\%$\downarrow$} & \cellcolor{mygray}MOTA$\uparrow$ & FP\%$\downarrow$ & Miss\%$\downarrow$ & \multicolumn{1}{c}{IDS\%$\downarrow$} & \cellcolor{mygray}MOTA$\uparrow$ & FP\%$\downarrow$ & Miss\%$\downarrow$ & IDS\%$\downarrow$ \\ \hline
AB3DMOT~\cite{abmot3d}                       & \cellcolor{mygray}40.1 & 16.4 & 43.4   & 0.13  & \cellcolor{mygray}37.7  & 11.6 & 50.2   & 0.47  & \cellcolor{mygray}-     & -    & -     & -      \\
PVRCNN-KF~\cite{shi2021pv}                  & \cellcolor{mygray}57.7 & 8.4  & 33.6   & 0.26  &\cellcolor{mygray} 53.8 & 9.3  & 36.2   & {0.73}  & \cellcolor{mygray}55.1  & 8.3  & 35.8  & 0.91   \\
AlphaTrack~\cite{zeng2021cross}                    & \cellcolor{mygray}55.7 & 9.6  & 34.3   & {0.44}  & \cellcolor{mygray}56.8  & 10.7 & 31.3   & {1.23}  & \cellcolor{mygray}59.6 & 5.4  & 33.7  & 1.23   \\
CenterPoint~\cite{yin2021center}                   & \cellcolor{mygray}59.4 & 9.4  & 30.9   & {0.32}  & \cellcolor{mygray}56.6  & 9.3  & 33.1   & {1.07}  & \cellcolor{mygray}60.0  & 11.1 & 28.1  & 0.78   \\
SimpleTrack~\cite{simpletrack}                   & \cellcolor{mygray}60.3 & 8.8  & 30.9   & {0.08}  & \cellcolor{mygray}60.1  & 10.7 & 28.8   & {0.40}  & \cellcolor{mygray}60.1  & 9.7  & 29.6  & 0.67   \\
ImmotralTrack~\cite{immortaltracker}                 & \cellcolor{mygray}60.6 & 8.5  & 31.0   & {0.01}  & \cellcolor{mygray}60.6  & 11.0 & 28.3   & {0.18}  & \cellcolor{mygray}61.6  & 9.3  & 29.0  & 0.10    \\ \hline
Ours (CenterPoint) & \cellcolor{mygray}\textbf{64.6}    & 8.5    & 26.7      & {0.17}     & \cellcolor{mygray}\textbf{62.3}     & 7.6    & 29.7      & {0.35}     & \cellcolor{mygray}\textbf{64.6}     & 8.7    & 26.1    & 0.64      \\ 
Ours (MPPNet) & \cellcolor{mygray}\textbf{64.9}    & 9.1    & 25.8      & {0.21}     & \cellcolor{mygray}\textbf{65.5}     & 9.4    & 24.7      & {0.42}     & \cellcolor{mygray}64.2    & 7.2    & 28.0    & 0.55      \\ \hline

\end{tabular}
\caption{Tracking performance on the Waymo Open dataset \textit{testing} split.}
\label{tab:waymo-test}
\vspace{-3mm}
\end{table*}

\noindent
\textbf{Implementation Details.}
We employ the detection boxes of CenterPoint and MPPNet as inputs for our method. 
During training, we take 4 hypotheses that includes 2 generated hypotheses (1 predicted box and 1 detection box) and 2 augmented hypotheses derived from the generated ones for diverse hypotheses distribution. For inference,
we specify the number of multiple hypotheses for each history trajectory as 6 (5 predicted boxes and 1 detection box) and 2 ( 1 predicted boxes and 1 detection box) for CenterPoint and MPPNet on Waymo, respectively. The detection boxes are associated with the trajectory through a greedy matching algorithm. For history trajectory without the matched current frame detection boxes, we pad all-zero boxes to create a hypothesis. 
We set a maximum matching distance of 2m, 0.5m and 1m for
vehicles, pedestrians and cyclists in Waymo and 2.2m, 2.0m for car and pedestrian in nuScenes, respectively.
The track-birth confidence threshold varies by detectors and classes. For CenterPoint, the track-birth confidence threshold is 0.2 for all classes in nuScenes and it is set to 0.72 for pedestrian and 0.8 for vehicle and cyclist in Waymo. For model hyper-parameters, we set the feature dimension $D=256$, the number of sampling points $Y=128$. 
We set the number of iteration blocks to 3 for both point feature encoding process and global-local interaction module. For optimization, the network is trained with the ADAM optimizer for 6 epochs with an initial learning rate of 0.001 and a batch size of 4.

\noindent
\textbf{Life Cycle Management.}
If the score of a trajectory's latest predicted hypothesis is below a threshold,
we remove the tracked object. For the retained objects, we select the hypothesis with the highest score as the association result.
Finally, the new-born objects are generated from detection boxes which remain unassociated with history trajectories and do not overlap with the history trajectories. 
These boxes that meet the criteria and have score above the track-birth threshold
are considered to new-born trajectories.

\subsection{Comparison with State-of-the-art 3D MOT Tracker}
\noindent
\textbf{Waymo Validation Set.}  In Table~\ref{tab:waymo-val}, we compare TrajectoryFormer with
other 3D MOT methods on the validation set of Waymo Open dataset, where TrajectoryFormer exhibits superior performance compared to other methods. To be specific, it outperforms the highest reported performance by 3.3$\%$, 0.5$\%$, and 1.5$\%$ and exceeds the adopted CenterPoint baseline by 4.6$\%$, 6.1$\%$, and 3.2$\%$ in terms of MOTA metric on vehicle, pedestrian, and cyclist, respectively.
More specifically, TrajectoryFormer exhibits a significant improvement in the Miss metric compared to the employed baseline, which implies that our method can successfully recover objects that were missed by the detector.  We attribute this success to our multi-hypothesis tracking strategy, which utilizes multiple trajectory hypotheses to propagate the state information of objects from past frames to the current frame and thus our model can better capture the potential motion of the tracked objects.
Besides, this strategy provides extra candidate bounding boxes, which can be associated with objects that the detector failed to detect in the current frame.
Moreover, for pedestrians, our method achieves lower False Positive (FP) values compared to other methods, indicating that the boxes in our trajectories have higher quality. Pedestrian trajectories are more complex and crowded compared to other categories, which makes it challenging for the network to generate correct associations. Hence, the lower FP for pedestrians indicates that TrajectoryFormer can handle associations in complex scenarios.
When adopt more advanced detector, MPPNet, TrajectoryFormer can achieve higher performance.

\noindent
\textbf{Waymo Testing Set.} 
As shown in Table~\ref{tab:waymo-test}, TrajectoryFormer also significantly outperforms other methods on the testing set of Waymo Open Dataset.

\noindent
\textbf{NuScenes Validation Set.} 
We also evaluate TrajectoryFormer on the validation split of the nuScenes dataset, as shown in Table~\ref{tab:nuscenes}. 
Following SpOT, we conduct experiments on the two main classes, namely car and pedestrian. All compared methods utilizes the detection results of CenterPoint. Our approach surpasses CenterPoint by 1.2\% and 5.6\% and SpOT by 0.3\% and 0.4\% in terms of AMOTA for car and pedestrian, respectively.

\begin{table}[]
\begin{center}
\scalebox{0.75}{
\begin{tabular}{lcccc}
\hline
\multirow{2}{*}{Method} & \multicolumn{2}{c}{Car} & \multicolumn{2}{c}{Pedestrian} \\
                        & \cellcolor{mygray}AMOTA$\uparrow$      & MOTA$\uparrow$      & \cellcolor{mygray}AMOTA$\uparrow$          & MOTA$\uparrow$         \\ \hline
CenterPoint     & \cellcolor{mygray}84.2          & 71.9          & \cellcolor{mygray}77.3       & 64.5             \\
SimpleTrack             & \cellcolor{mygray}83.8          & 70.1          & \cellcolor{mygray}79.4       & 67.0             \\
ImmotralTracker         & \cellcolor{mygray}84.0          & 69.8         & \cellcolor{mygray}80.2       & 68.0             \\
SpOT         & \cellcolor{mygray}85.1          & -         & \cellcolor{mygray}82.5       & -             \\ \hline
TrajectoryFormer (ours)       & \cellcolor{mygray}\textbf{85.4}          & \textbf{75.0}          &\cellcolor{mygray}\textbf{82.9}       &  \textbf{69.9}            \\ \hline
\end{tabular}
}
\end{center}
\vspace{-0.3cm}
\caption{Tracking performance on \emph{val} split of the nuScenes dataset. All the compared methods utilize CenterPoint as the detector, and the main metric (AMOTA) is highlighted in gray.}
\label{tab:nuscenes}
\vspace{-0.5cm}
\end{table}

\subsection{Ablation Studies}
To verify the effectiveness of each component in TrajectoryFormer, We conduct comprehensive ablation studies on the Waymo benchmark. Unless otherwise mentioned, all ablation experiments of TrajectoryFormer are trained on the vehicle category by taking the detection results of CenterPoint with 3 epoch. We take MOTA (LEVEL 2) as the default metric for comparison.

\noindent
\textbf{Effects of the multiple hypotheses.}
Table~\ref{tab:num_hypo} investigates the impact of different number of hypotheses for each tracked object.
Firstly, without predicted boxes, TrajectoryFormer's association performance degrades to the same level as CenterPoint baseline, which employs center distance and a greedy algorithm to perform trajectory-box association for each trajectory. 
In this scenario, compared to baseline, the refinement of detection box results in a 1.2$\%$ performance gain.
For the single prediction box setting, we set $T_f = 1$. In other words, the motion prediction network only predicts the future box of the tracked object in the next single-frame. 
The incorporation of even a single prediction boxes allows the network to transfer past information of tracked objects to the current frame, resulting in a significant 3.5$\%$ performance improvement.
When employing multiple temporal prediction boxes (\emph{e.g.}, 5), a slight performance improvement of 0.3$\%$ is observed compared to the single-frame prediction box setting. 
Utilizing the prediction from trajectory embedding at different history moments can provide more diverse candidate boxes, which brings a slight improvement. However, the use of more prediction boxes (\emph{i.e.}, 10) does not provide any additional performance improvements and instead increases computational overhead. 

\begin{table}[]
\centering
\scalebox{0.9}{
\begin{tabular}{lcccc}
\hline
Method             & \cellcolor{mygray}MOTA$\uparrow$ & FP$\downarrow$  & Miss$\downarrow$ & IDS$\downarrow$ \\ \hline
CenterPoint~\cite{yin2021center}  & \cellcolor{mygray}55.1 & 10.8 &33.9 &0.26 \\
w/o pred. boxes             & \cellcolor{mygray}56.3 &10.5  & 33.0 & 0.24 \\
1 pred. box            & \cellcolor{mygray}59.5 & 12.1 & 28.3 & 0.21 \\
5 pred. boxes             & \cellcolor{mygray}\textbf{59.8} & 11.3 & 28.7 & 0.23 \\
10 pred. boxes            & \cellcolor{mygray}59.7 & 11.6 & 28.5 & 0.22 \\ \hline
\end{tabular}
}
\vspace{1mm}
\caption{Effects of the number of temporally prediction boxes. All experiments use 1 heuristic matched detection box.}
\label{tab:num_hypo}
\end{table}

\begin{table}
\centering
\scalebox{0.9}{
\begin{tabular}{cccccc}
\hline
Category & Method  & \cellcolor{mygray}MOTA$\uparrow$ & FP$\downarrow$  & Miss$\downarrow$ & IDS$\downarrow$  \\ \hline
\multirow{3}{*}{Vehicle} &
1 frame & \cellcolor{mygray}59.6   & 11.5   & 28.7   & 0.23 \\
& 3 frame & \cellcolor{mygray}\textbf{59.8}   & 11.3   & 28.7   & 0.23 \\
& 5 frame & \cellcolor{mygray}\textbf{59.8} & 11.3 & 28.7 & 0.23 \\ \hline
        \multirow{3}{*}{Pedestrian} &
1 frame & \cellcolor{mygray}59.8      & 9.7     & 30.1      & 0.37   \\
& 3 frame & \cellcolor{mygray}60.8     & 8.9     & 29.9      & 0.37    \\
& 5 frame & \cellcolor{mygray}\textbf{61.0}    & 8.8    & 29.8  & 0.37  \\ \hline
\end{tabular}
}
\vspace{1mm}
\caption{Effects of different numbers of point cloud frames for appearance feature encoding.}
\label{tab:num_point_frame}
\vspace{-5mm}
\end{table}

\noindent
\textbf{Effects of length of point cloud frames.}
Table~\ref{tab:num_point_frame} displays the performances of using different numbers of point cloud frames. It should be noted that, in contrast to SpOT~\cite{stearns2022spot}, which maintains a point cloud sequence with the same length as the trajectory bounding box, we only crop the concatenated multi-frame points of hypothesis boxes at the current time to reduce computation overhead. we scrutinize the effect of different numbers of point cloud frames in appearance feature encoding by keeping the number of randomly sampled point clouds constant. For the vehicle class, the point cloud appearance information of 1, 3, or 5 frames yields comparable performance. Conversely, for the pedestrian class, the utilization of 5-frame point cloud information outperforms single-frame and 3-frame point clouds by $1.2\%$ and $0.2\%$, respectively. We attribute this to the fact that the pedestrian class has sparser raw LiDAR points in comparison to vehicles. Thus, the concatenated multi-frame points can provide more complete appearance information, which is advantageous for the network to differentiate between various candidate hypotheses.

\noindent
\textbf{Effects of length of trajectory boxes.}
we explore the impact of trajectory box length of our approach, as presented in Table \ref{tab:num_traj_length}. We observe that trajectories that are too short fail to fully leverage past temporal motion information of the tracked objects, resulting in $0.5\%$ performance drop. For the Waymo dataset, we find that history trajectory boxes length with 10 frames can effectively capture the object's past motion states, resulting in the best performance. Further increasing the trajectory length does not yield any additional performance benefits, as the motion state of the object may have changed, compared with earlier time steps. However, longer trajectories result in additional computational overhead. Therefore, we employ 10 frame trajectory boxes as the default setting in our approach as a trade-off.

\noindent
\textbf{Effects of the combination of point embedding and trajectory embedding.}
Table~\ref{tab:embed} presents the investigation of different hypothesis embedding designs. As we can see, only using the long-term boxes feature will lead to a 9$\%$ performance drop, which is reflected by the large value of the Miss and FP indicators. This suggests that a network based solely on the trajectory boxes feature cannot adequately select the best matching boxes for each tracked object, resulting in the retention of low-quality boxes (increasing FP) and the discarding of high-quality boxes (increasing Miss). Meanwhile, utilizing the short-term appearance features of point clouds demonstrates better association ability than trajectory box features, but also decreases performance by 3.3$\%$. In the end, the optimal performance was achieved through the joint utilization of point cloud and trajectory features, emphasizing the significance of integrating both motion and appearance information.

\begin{table}[]
\centering
\scalebox{0.9}{
\begin{tabular}{ccccc}
\hline
Method             & \cellcolor{mygray}MOTA$\uparrow$ & FP$\downarrow$  & Miss$\downarrow$ & IDS$\downarrow$ \\ \hline
5 frame             &\cellcolor{mygray} 59.3 & 11.7 & 28.8 & 0.23 \\
10 frame            & \cellcolor{mygray}\textbf{59.8} & 11.3 & 28.7 & 0.23 \\
15 frame            & \cellcolor{mygray}59.7 & 11.5 & 28.6 & 0.22 \\ 
20 frame            & \cellcolor{mygray}59.7 & 11.5 & 28.6 & 0.23 \\\hline
\end{tabular}}
\vspace{1mm}
\caption{Effects of numbers of trajectory length.}
\label{tab:num_traj_length}
 \vspace{-3mm}
\end{table}

\begin{table}[]
\centering
\scalebox{0.9}{
\begin{tabular}{lcccc}
\hline
Method             & \cellcolor{mygray}MOTA$\uparrow$ & FP$\downarrow$  & Miss$\downarrow$ & IDS$\downarrow$ \\ \hline
Trajectory         & \cellcolor{mygray}50.8 & 15.4 & 33.2 & 0.59 \\
Point              & \cellcolor{mygray}56.5 & 12.2 & 31.2 & 0.17 \\
Point + Trajectory & \cellcolor{mygray}\textbf{59.8} & 11.3 & 28.7 & 0.23 \\ \hline
\end{tabular}}
\vspace{1mm}
\caption{Effects of different designs of hypothesis embedding.}
\label{tab:embed}
\vspace{-6mm}
\end{table}

\section{Conclusion}
In conclusion, our work presents a novel transformer-based 3D MOT tracking framework, TrajectoryFormer, which immigrates the limitations of existing tracking-by-detection methods by leveraging multiple predictive hypotheses that incorporate both temporally predicted boxes and current-frame detection boxes. To better encode spatial-temporal information of each hypothesis with low computational overhead, we incorporate both long-term trajectory motion features and short-term point appearance features. Additionally, our global-local interaction module enables the exploitation of context information by modeling relationships among all hypotheses. Extensive experiments on the Waymo 3D tracking benchmark demonstrate that our proposed approach outperforms existing state-of-the-art methods, validating the effectiveness of our framework.

\section{Acknowledgement}
This project is funded in part by National Key R\&D Program of China Project 2022ZD0161100, by the Centre for Perceptual and Interactive Intelligence (CPII) Ltd under the Innovation and Technology Commission (ITC)’s InnoHK, by General Research Fund of Hong Kong RGC Project 14204021. Hongsheng Li is a PI of CPII under the InnoHK.

{\small
\bibliographystyle{ieee_fullname}
\bibliography{egbib}
}

\end{document}